\documentclass{article}

\usepackage{arxiv}

\usepackage[utf8]{inputenc} 
\usepackage[T1]{fontenc}    
\usepackage{hyperref}       
\usepackage{url}            
\usepackage{booktabs}       
\usepackage{amsfonts}       
\usepackage{nicefrac}       
\usepackage{microtype}      
\usepackage{lipsum}	
\usepackage{graphicx}
\usepackage{doi}
\usepackage[export]{adjustbox}

\title{AMDet: A Tool for Mitotic Cell Detection in Histopathology Slides}

\author{Walt Williams\thanks{Walt Williams is a student at the University of Memphis and was an intern at Microsoft Research when this work was done.} \\
	AutoML Image Lab\\
	Microsoft Research\\
	Cambridge, MA 02142 \\
	\texttt{t-wwilliams@microsoft.com} \\
	\And
    James Hall \\
	AutoML Image Lab\\
	Microsoft Research\\
	Cambridge, MA 02142 \\
	\texttt{James.Hall@microsoft.com}\\}

\renewcommand{\headeright}{Technical Report}
\renewcommand{\undertitle}{Technical Report}
\renewcommand{\shorttitle}{AMDet: A Tool for Mitotic Cell Detection in Histopathology Slides}

\hypersetup{
pdftitle={A template for the arxiv style},
pdfsubject={q-bio.NC, q-bio.QM},
pdfauthor={David S.~Hippocampus, Elias D.~Striatum},
pdfkeywords={First keyword, Second keyword, More},
}

\begin{document}
\maketitle

\begin{abstract}
Breast Cancer is the most prevalent cancer in the world. The World Health Organization reports that the disease still affects a significant portion of the developing world citing increased mortality rates in the majority of low to middle income countries. The most popular protocol pathologists use for diagnosing breast cancer is the Nottingham grading system which grades the proliferation of tumors based on 3 major criteria, the most important of them being mitotic cell count. The way in which pathologists evaluate mitotic cell count is to subjectively and qualitatively analyze cells present in stained slides of tissue and make a decision on its mitotic state i.e. is it mitotic or not? This process is extremely inefficient and tiring for pathologists and so an efficient, accurate, and fully automated tool to aid with the diagnosis is extremely desirable. Fortunately, creating such a tool is made significantly easier with the AutoML tool available from Microsoft Azure, however to the best of our knowledge the AutoML tool has never been formally evaluated for use in mitotic cell detection in histopathology images. This paper serves as an evaluation of the AutoML tool for this purpose and will provide a first look on how the tool handles this challenging problem. All code is available at \href{https://github.com/WaltAFWilliams/AMDet}{\underline{https://github.com/WaltAFWilliams/AMDet}}.
\end{abstract}


\section{Introduction}
Breast Cancer is the most prevalent cancer in the world and represents roughly 25\% of all cancers diagnosed in women \cite{WHOStats}. The CDC reports that the annual number of new breast cancer diagnoses in the United States has risen every year since 2011, and the trend is projected to sustain into the future \cite{CDCStats}. Even with these disparities, the WHO has reported that current Breast Cancer treatments are highly effective at dealing with the cancer if spotted early, up to a 90\% 5-year survival rate. One of the most common methods pathologists use to diagnose breast cancer is analyzing histopathology slides of breast tissue that have been stained with Hematoxylin and Eosin (H\&E), 2 chemicals used to highlight structures within a cell \cite{AIMIT}. When examining these H\&E stained images the Nottingham grading system is often used to grade the aggressiveness of the cancer. In this system there are three primary markers used for grading: nuclear atypia, tubule formation, and the mitotic cell count. The mitotic cell count is the most important marker among them because it is directly related to the prognosis of tumors \cite{AIMIT}.

The analysis of histopathology slides by pathologists is often a slow, tiring, and extremely subjective process susceptible to high variability among different pathologists. To aid in this task several competitions have been conducted with the intention of developing tools capable of aiding pathologists with the analysis of histopathology slides. Deep Learning has emerged as a successful method for creating these tools seeing as the top winners for one of the prominent competitions, the ICPR 2014 competition, was created with the use of deep learning \cite{ICPR2014}. In this paper we propose AutoML Mitosis Detector, or AMDet, a deep learning tool used to aid in the diagnosis of breast cancer from h\&e stained images by drawing bounding boxes around non-mitotic and mitotic cells in the images. AMDet works by taking in a single histopathology slide, dividing the image into patches, running  forward pass for each of the patches to aggregate predictions, then combining the patches back together to form the original slide and saving it to a specified directory. Figure 1 shows some examples of AMDet's predictions.

\begin{figure}
    \centering
    \includegraphics[width=4cm, height=4cm]{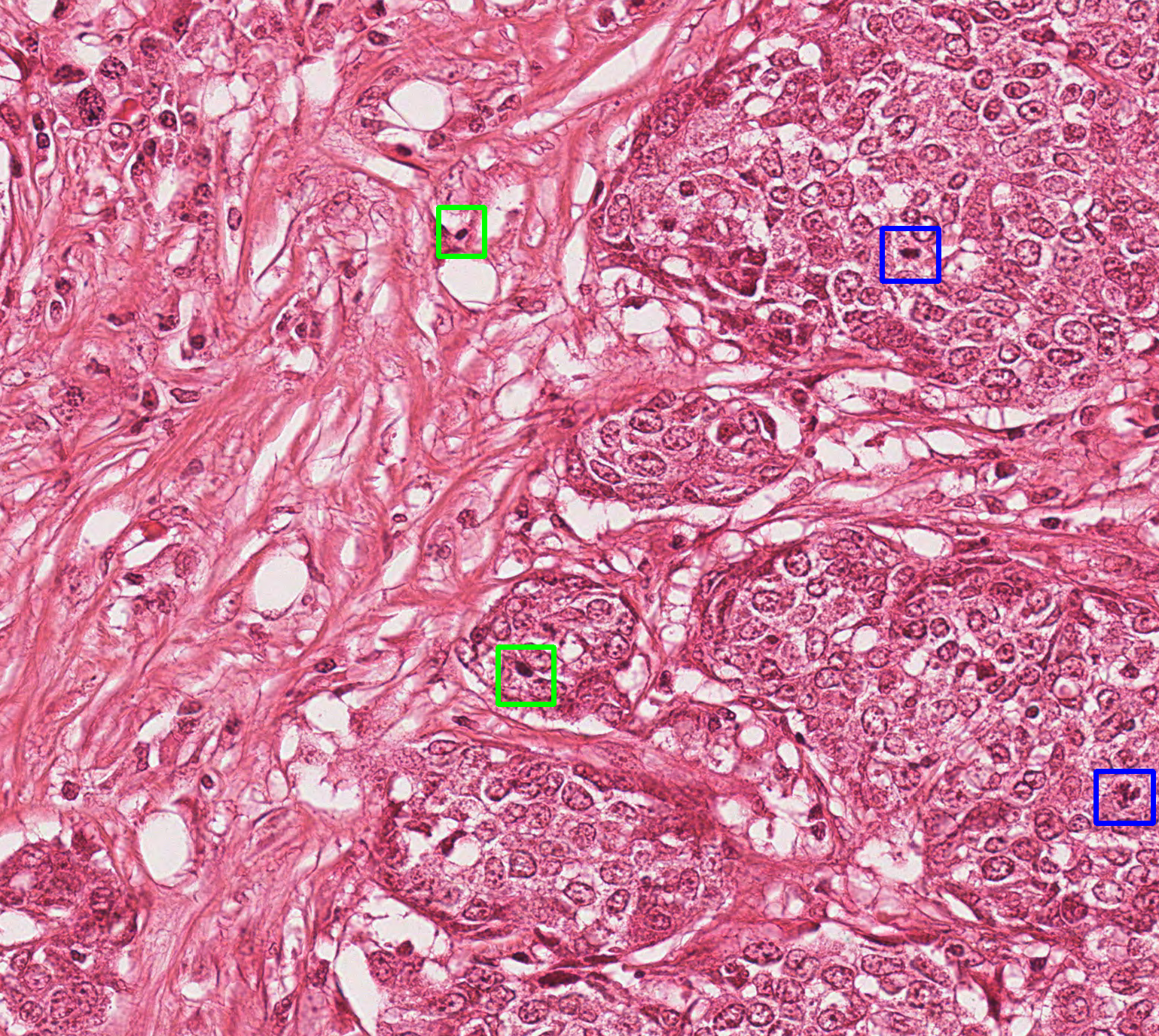}
    \includegraphics[width=4cm, height=4cm]{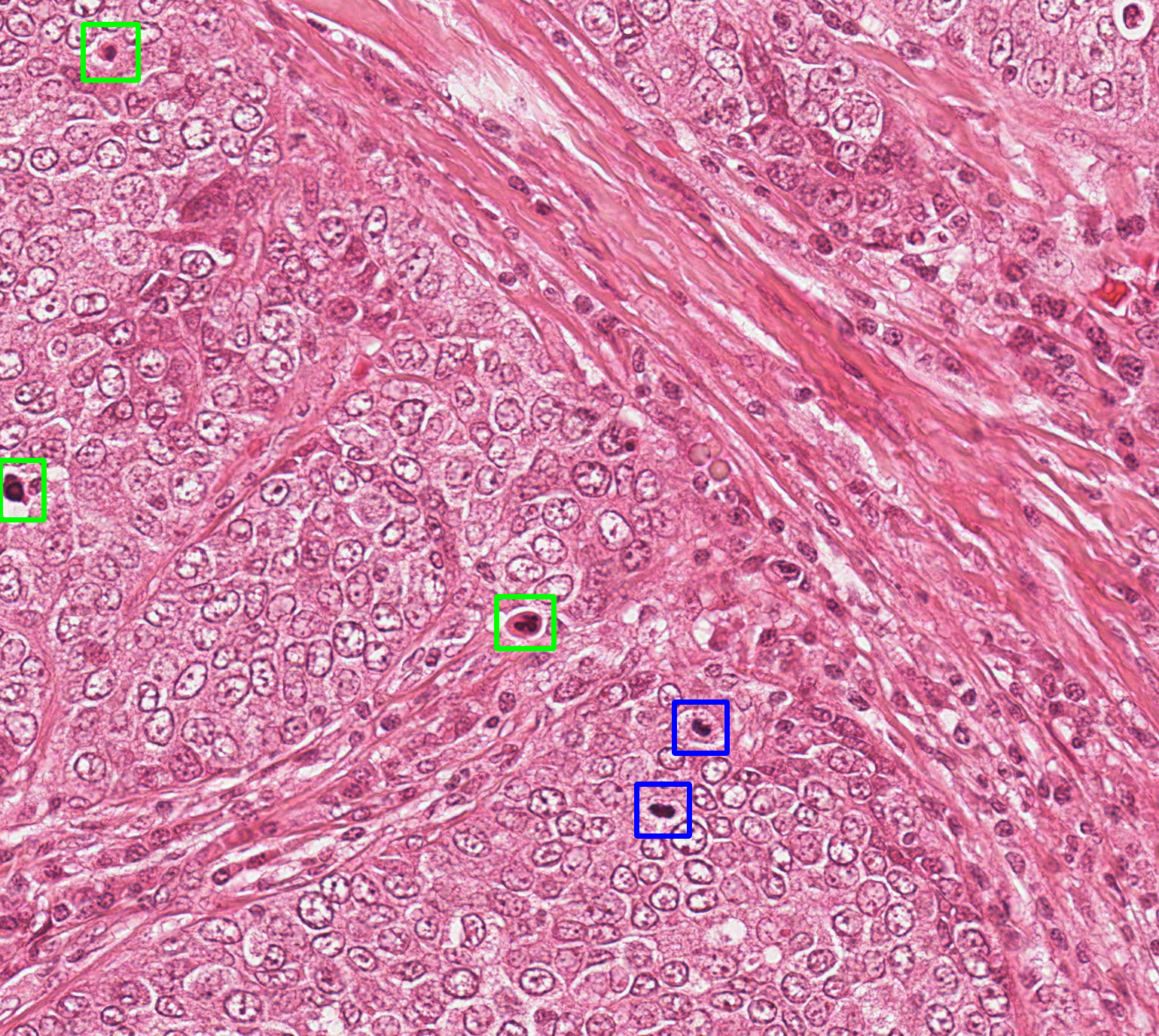}
    \includegraphics[width=4cm, height=4cm]{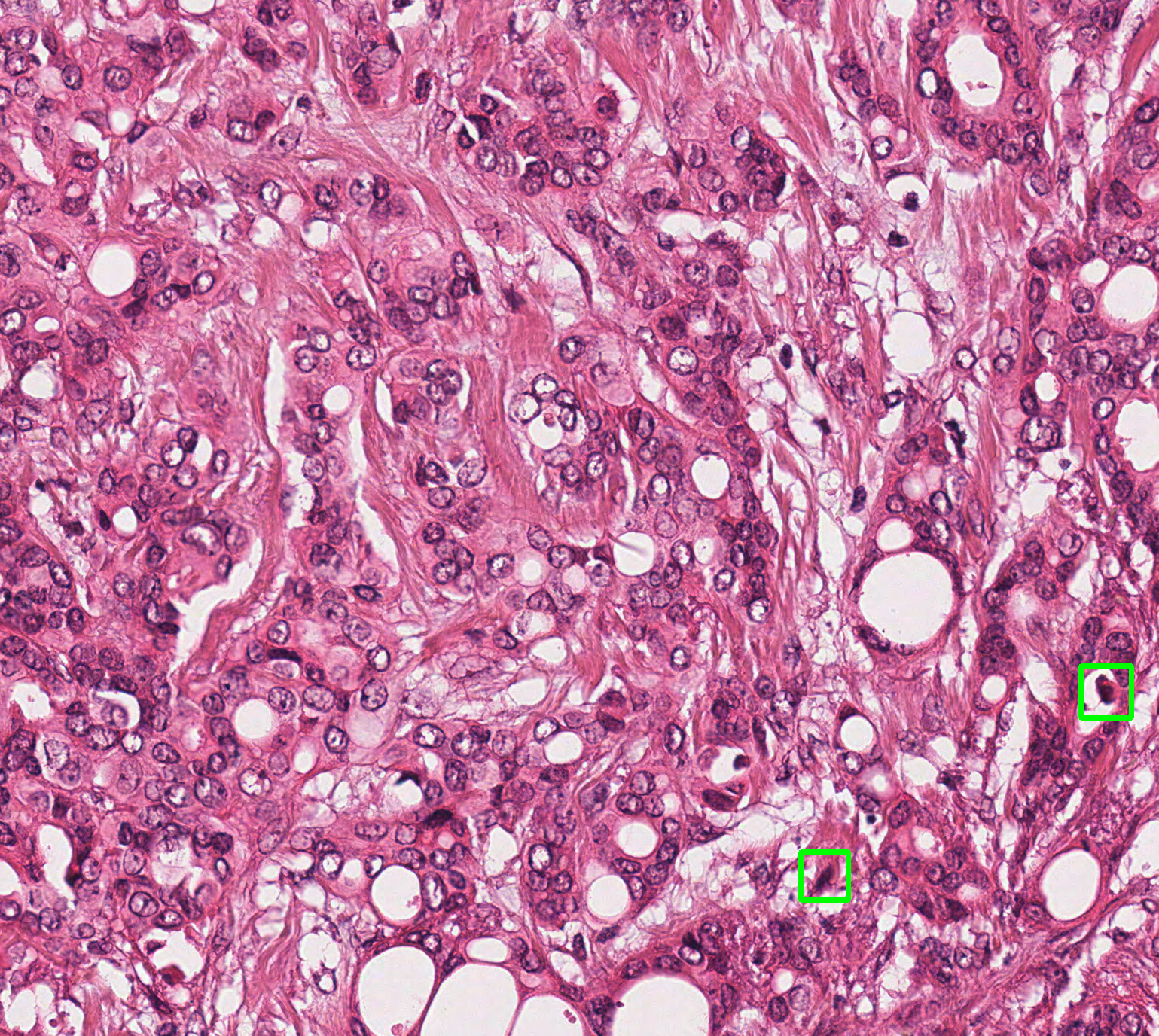}
    \includegraphics[width=4cm, height=4cm]{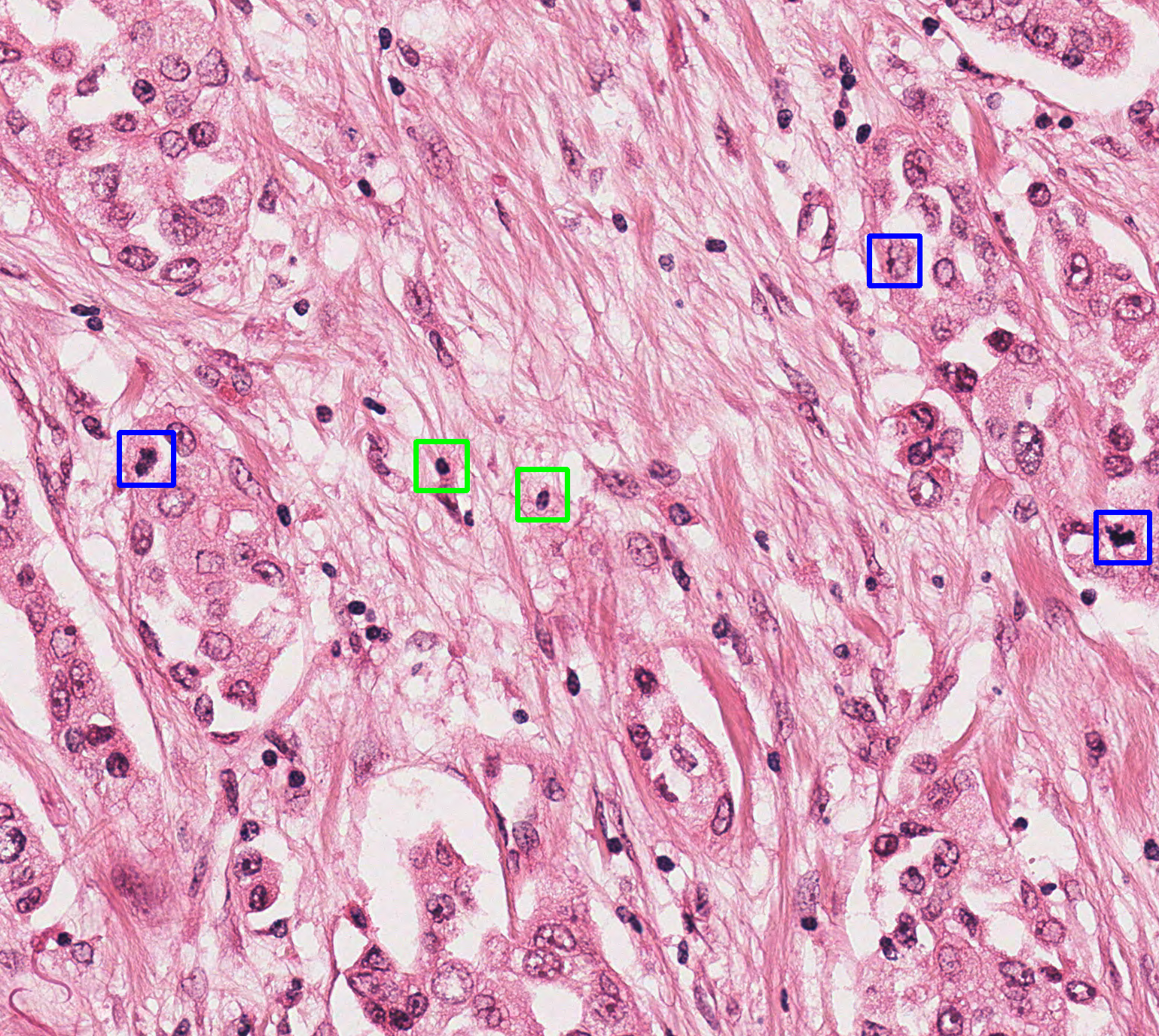}
    \caption{Example detections using AMDet on images taken from the ICPR2014 Dataset. Blue boxes indicate mitotic cells and green boxes are non-mitotic cells.}
    \label{fig:fig1}
\end{figure}

The goal of this paper is to serve as an evaluation of the AutoML tool for the purpose of detecting small objects (cells) present in medical images. The rest of the paper is divided as followed: Section 2 will be a literature review discussing recently proposed pipelines for the detection of mitotic figures in histopathology slides. Section 3 will detail the specific methodology used in the creation and evaluation of our pipeline. Section 4 discusses the results of the experiments. Section 5 is a discussion on limitations of this work and directions for future research. Section 6 is the conclusion. 

\section{Related Work}
\label{sec:relatedwork}
Machine Learning has seen a surge in popularity for the analysis of histopathology slides in recent years \cite{nuset, unet, hovernet, yancy}. Several models have been proposed in the literature for performing object detection, instance segmentation, and semantic segmentation in these slides with the aim to assist pathologists with their diagnoses. One of the most popular models used for semantic segmentation in medical images is the U-Net model. U-Net is a fully convolutional network (FCN) that features a down-sampling  path that plays the role of a feature extractor. The low-dimensional feature map is then sent through an upsampling path that combines the low-dimensional feature maps with its corresponding high-dimensional representation in the down-sampling branch. The result is a segmentation map that classifies each pixel in the image as belonging to a particular class \cite{unet}. It is a landmark study in the field of biomedical image segmentation and widely used. Another popular model proposed for nucleic segmentation is NuSet, short for Nuclear Segmentation Tool \cite{nuset}. NuSet is a deep learning model used to segment crowded and overlapping cells in histopathology images. It does so with three major components, a region proposal network to generate potential regions-of-interest inside an image, a U-Net to provide segmentation maps, and a watershed algorithm that approximates the borders of touching cells in order to segment them. Graham et al. propose HoVer-Net, a model used for instance segmentation and classification of cells located in histopathology images \cite{hovernet}. Yancy proposes Multi-Stream Faster-RCNN, a model for detecting bounding boxes around mitotic cells. This model combines a segmentation map from a U-Net with an RGB image into 2 separate Faster-RCNN streams in order to add more context to the detections \cite{yancy}.

\section{Methods}
\label{sec:methods}

\subsection{Network Architecture}
AMDet is created from a Faster-RCNN model using a pre-trained residual network with 50 layers (resnet-50) trained as its feature extractor. Faster-RCNN is a deep learning model introduced by Ren et al. for the purpose of object detection in natural images. It involves 3 major components: a Region Proposal Network to locate sections of the image which may contain an object (region proposals), followed by region-of-interest (ROI) pooling layers, and then classification+regression layers to predict the bounding box and classify the object inside \cite{ren}.

\begin{figure}
    \centering
    \includegraphics[width=2cm, height=2cm]{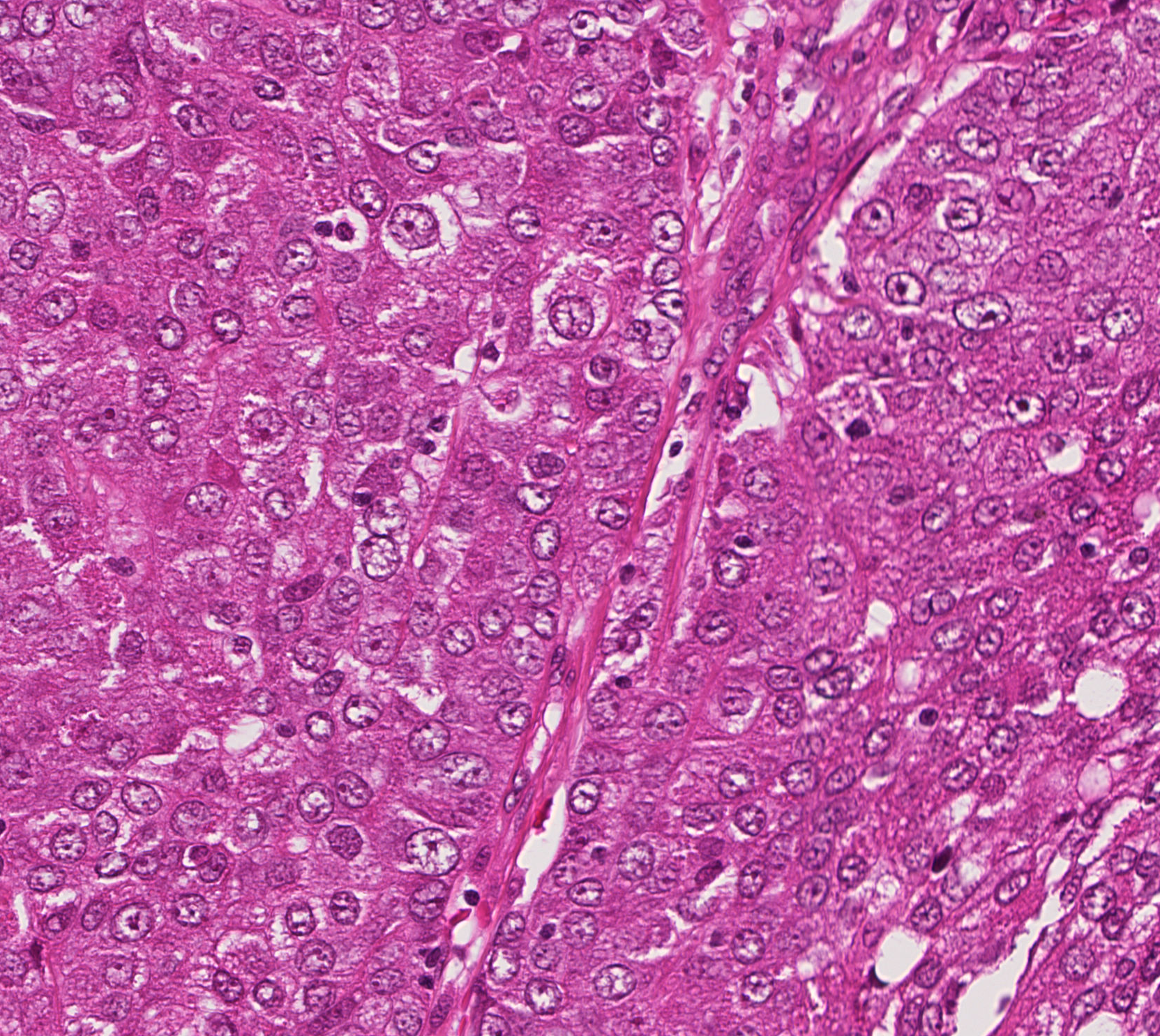}\\
    \includegraphics[width=2cm, height=2cm]{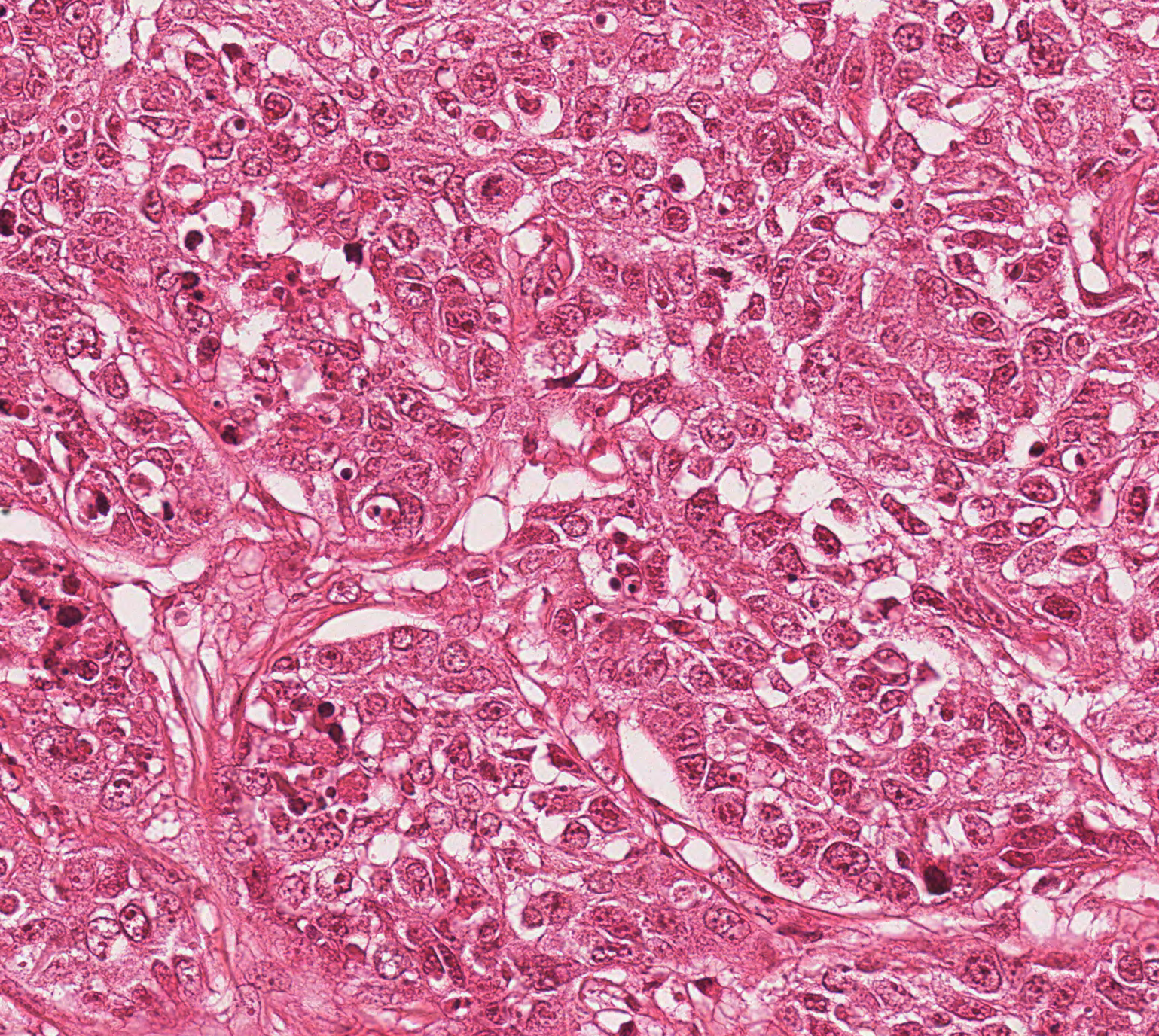}    \includegraphics[width=2cm, height=2cm]{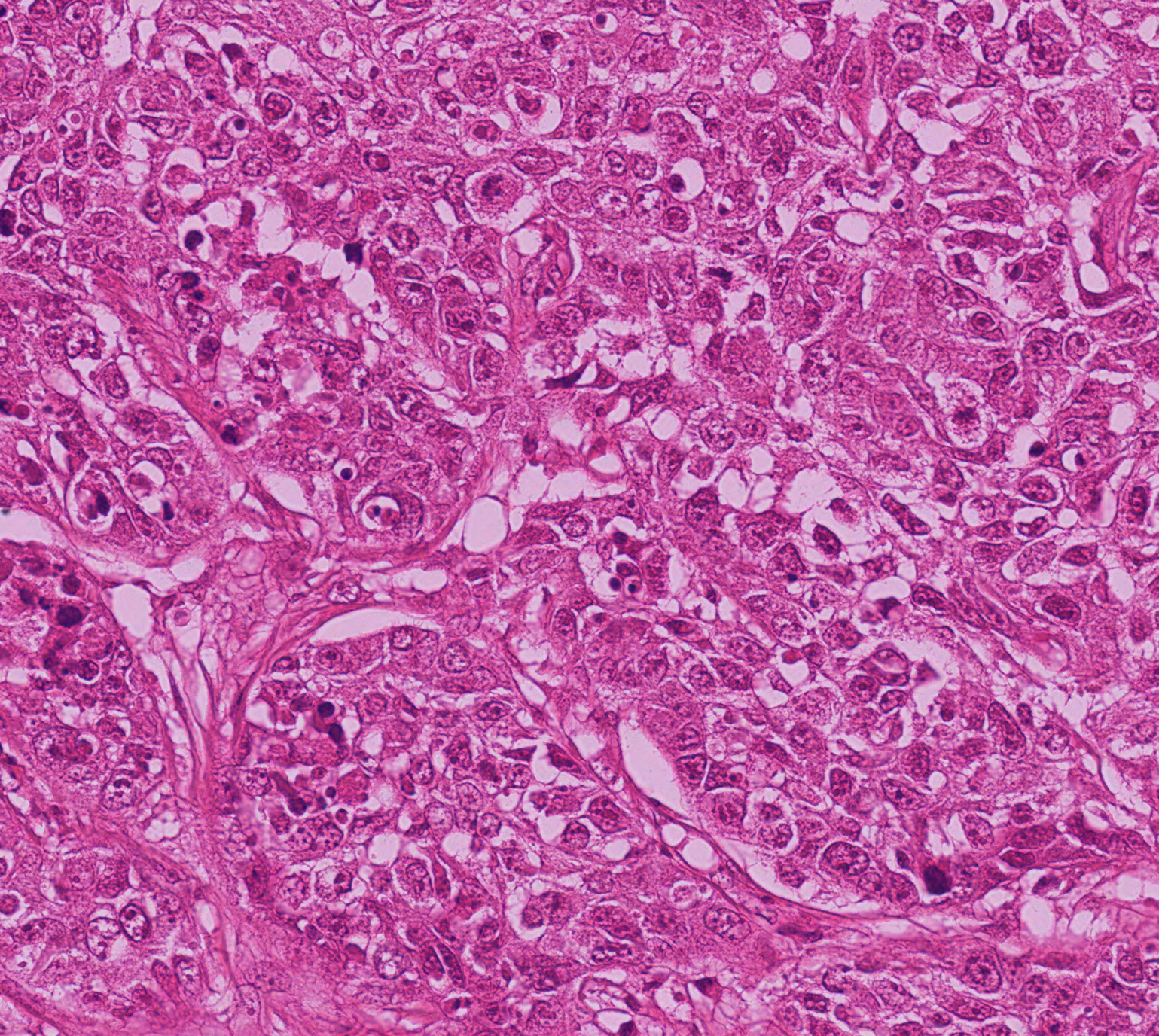}\\
    \includegraphics[width=2cm, height=2cm]{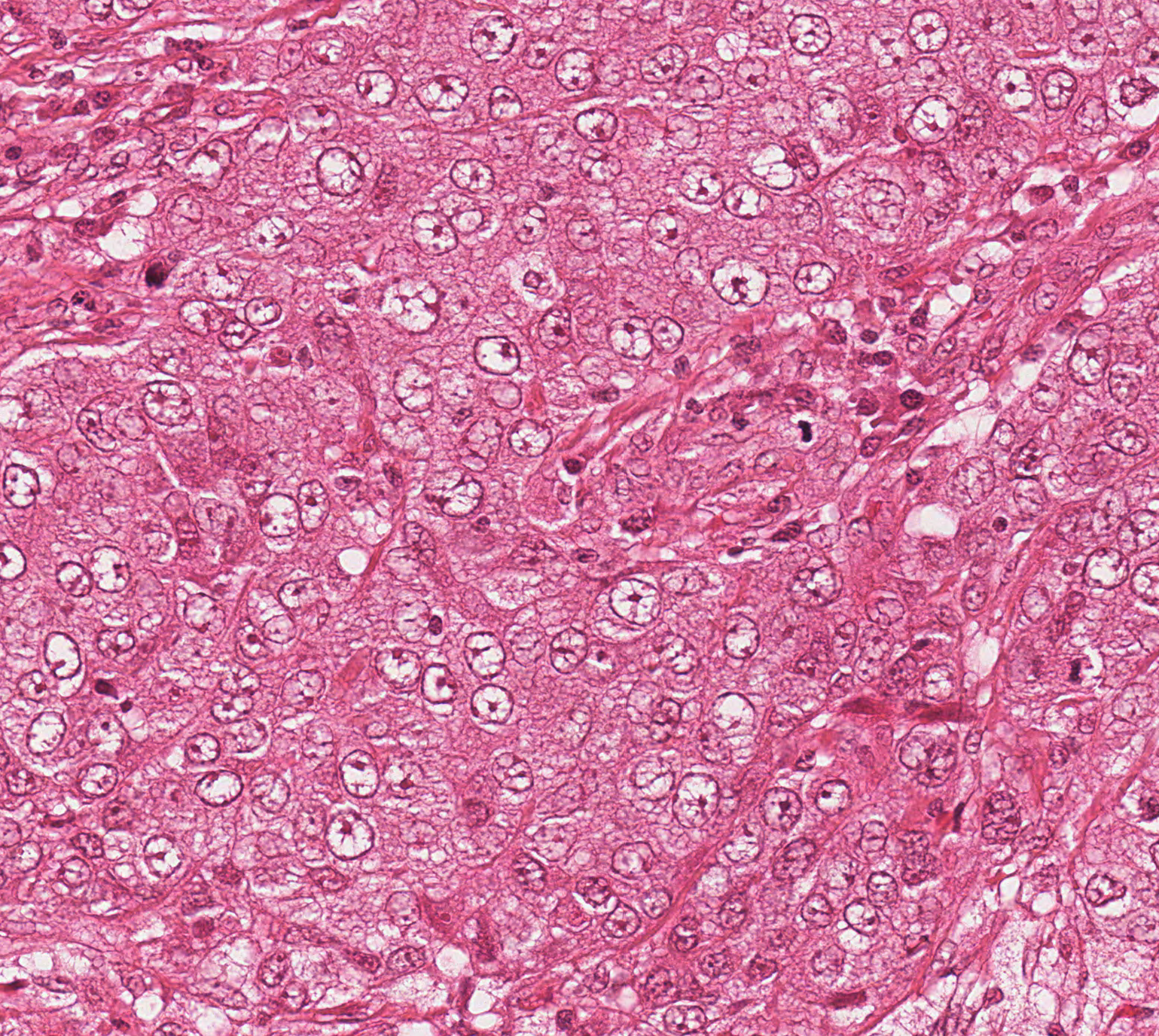}
    \includegraphics[width=2cm, height=2cm]{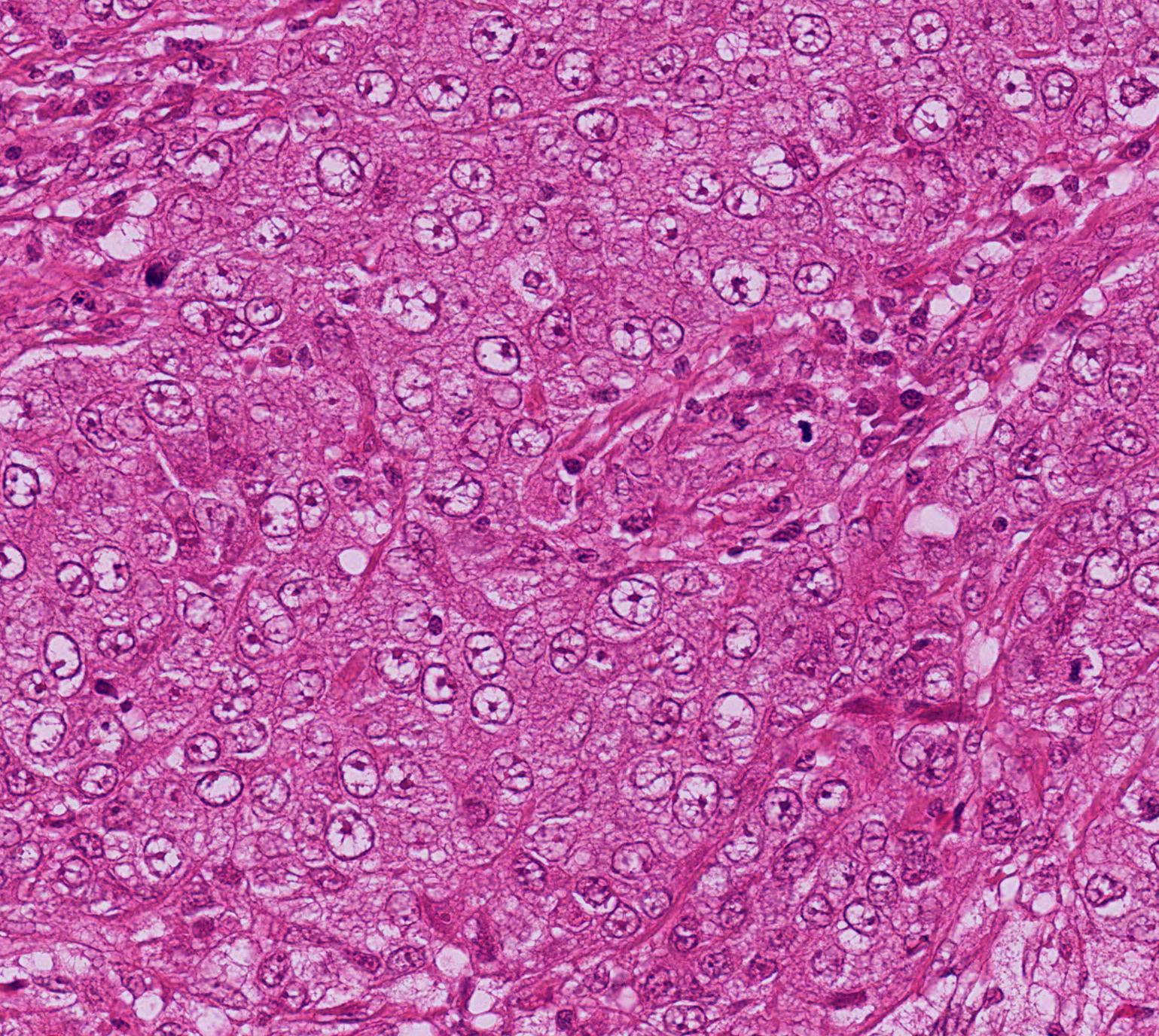}\\
    \includegraphics[width=2cm, height=2cm]{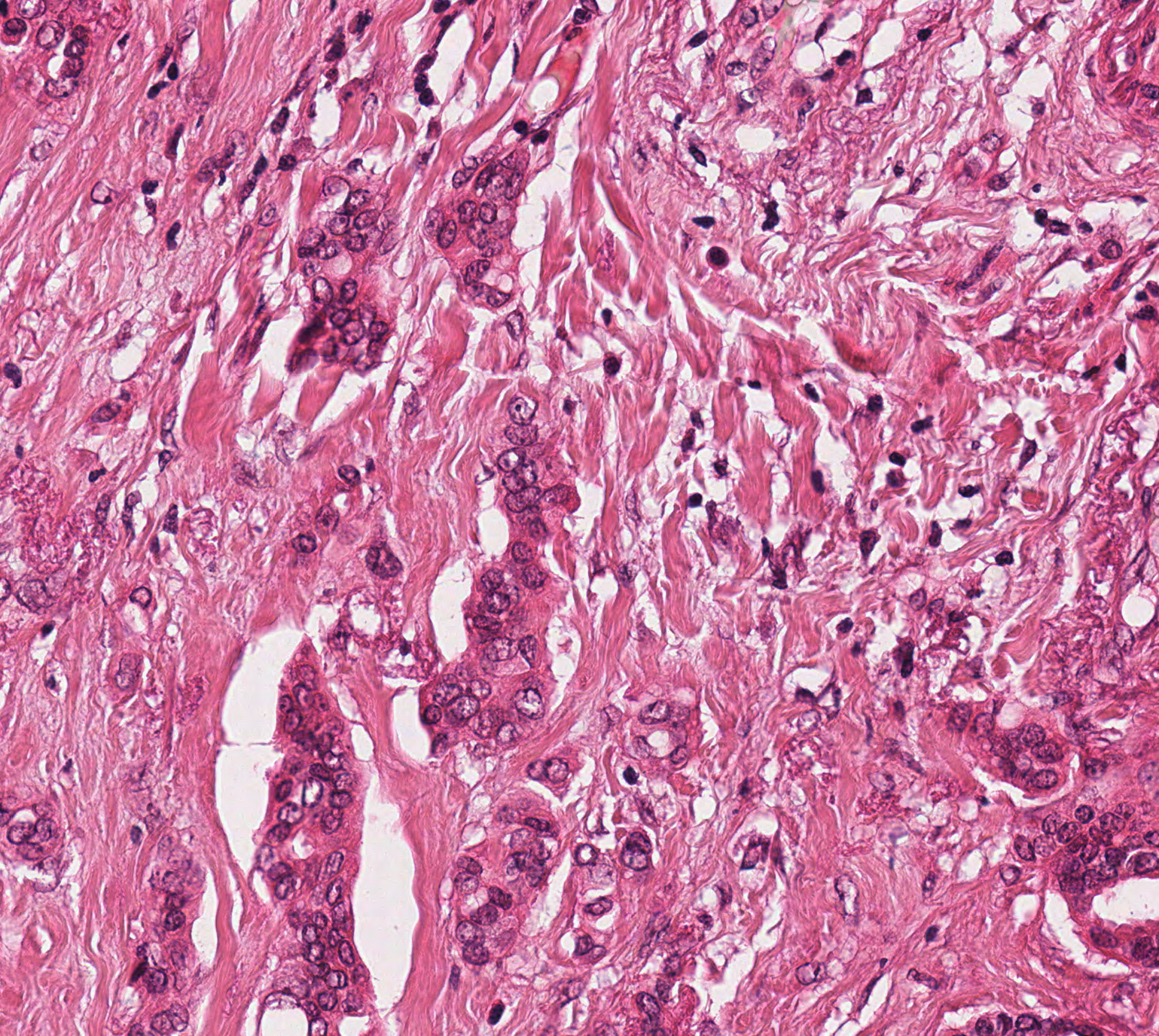}
    \includegraphics[width=2cm, height=2cm]{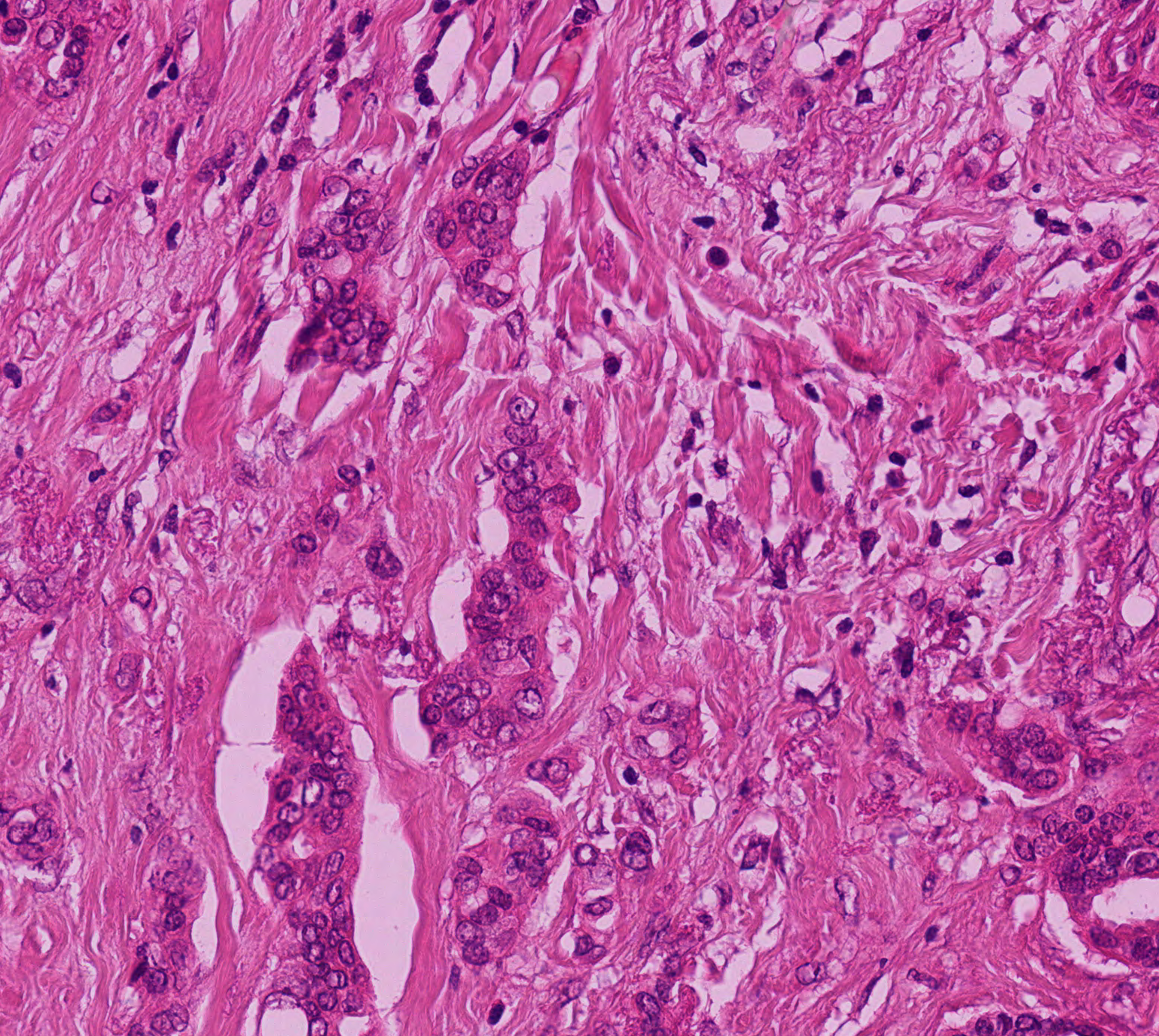}\\
    \includegraphics[width=2cm, height=2cm]{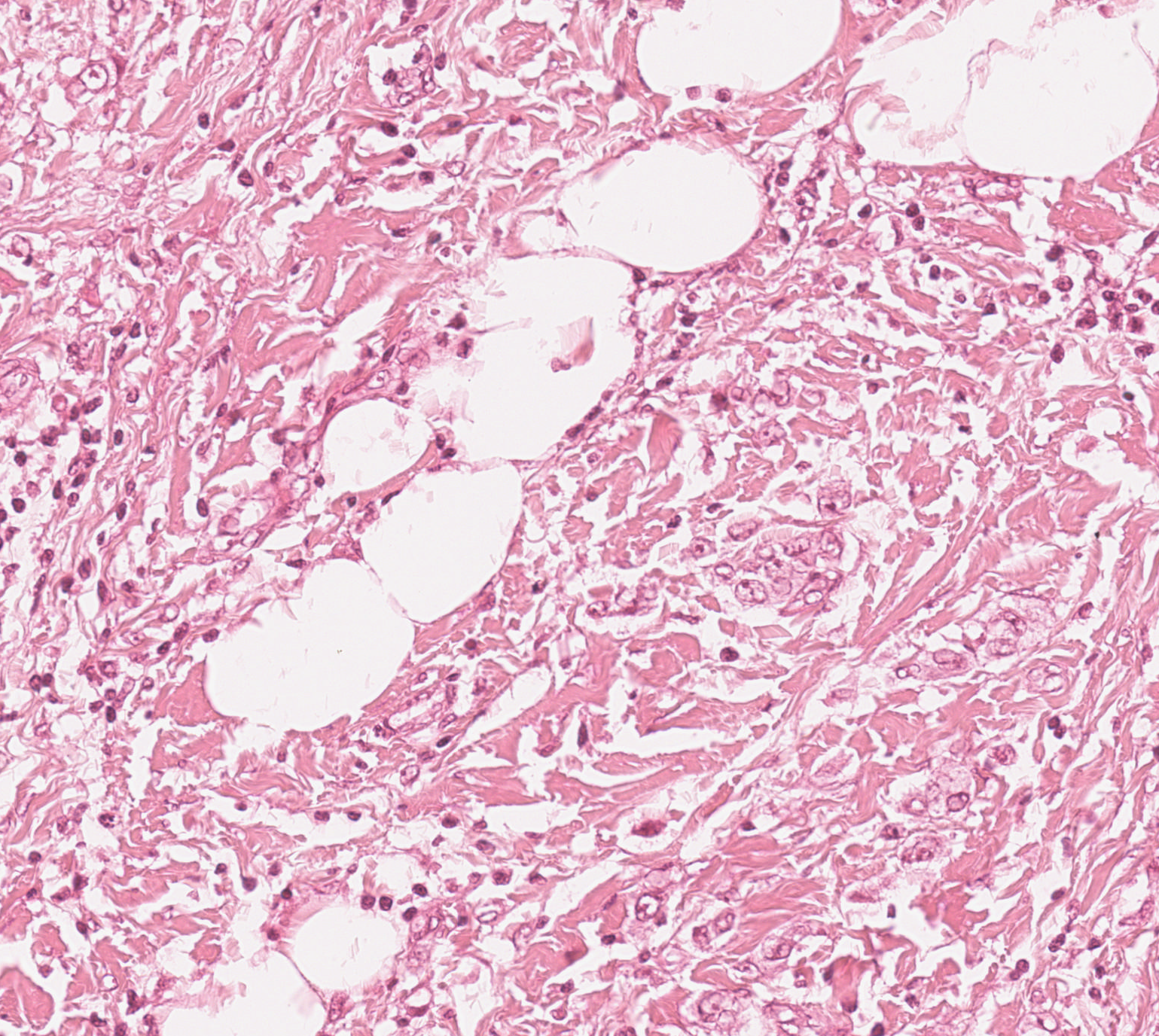}
    \includegraphics[width=2cm, height=2cm]{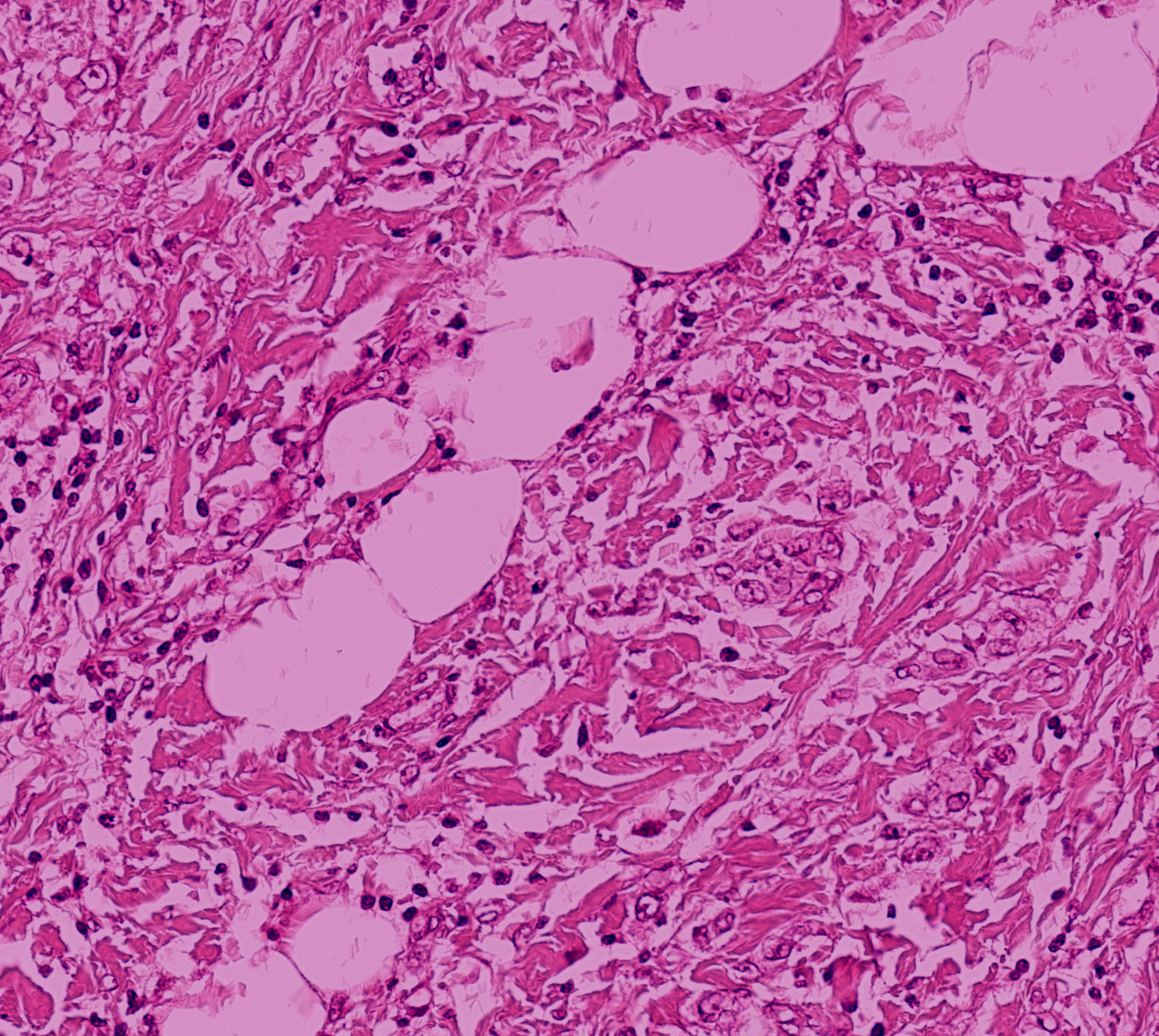}
    \caption{A set of images before and after being tranformed via Reinhard Normalization. The topmost image is the target which we will be trying to match in the rest of the images. The next 4 rows contain the original image followed by the image after normalization.}
    \label{fig:fig2}
\end{figure}

Faster-RCNN builds on previous work that proposed Regional-CNNs (R-CNNs) for the simultaneous localization and classification of objects inside images, also known as object detection \cite{girshickR, girshickF}. One issue with the original R-CNN model was the time it took to generate predictions. This was largely a  result of 2 main factors, a multi-stage pipeline leading to issues of repeated computations for all region proposals and the the use of selective search: an algorithm for proposing potential objects in an image in a heuristic manner. The issue of the multi-stage pipeline was largely alleviated with the work from the second version of R-CNN, called Fast-RCNN \cite{girshickF}. Fast-RCNN replaced the 3-stage detection pipeline of the original R-CNN with a unified framework trained end-to-end with a multi-task loss. After this change, the last bottleneck in terms of computational time was the region proposal algorithm being used by Fast-RCNN, selective search. To help speed up the process Faster-RCNN introduces a Region Proposal Network (RPN) in order to extract candidate regions from an image and then share weights with the rest of the network thereby replacing the old and slow selective search algorithm. The RPN is a modified CNN designed to extract features from the original image and generate region proposals to be sent to the ROI pooling layers. It does this with the use of anchors, which are boxes of varying aspect ratio and size that are placed on the original image at a location relative to their point on the output feature map. These anchors are then graded by their likelihood of containing an object before being sent to the ROI pooling layers. 

Once the candidate regions have been proposed, the next step is to prepare the ROIs for the pass into the fully connected dense layers for classification and regression. Dense layers depend on fixed-size feature vectors so the ROIs which come in varying shapes have to be resized to a fixed sized in order to be passed through. The ROI pooling layers resize the images by applying max pooling to the ROIs in order to reshape them into the correct size. When the images are correctly resized they are then sent to the final classification and regression layers to output the class and the bounding box coordinates for the object in the original image. 

\subsection{ICPR 2014 Mitosis Atypia}
AMDet was trained using h\&e stained slides collected from breast cancer tissue from the Mitos-Atypia-14 Grand challenge. The goal of the competition was to to create a model to classify instances of mitosis inside the slides based on morphological features. The original whole slide images were scanned with one of two slide scanners, the Aperio Scanscope XT and the Hamamatsu Nanozoomer 2.0-Ht. Pathologists then filtered out much of the unneeded information by only frames from the whole slide image that were located inside tumors. Afterwards the frames were subdivided into 4 slides at X40 magnification for grading of nuclear atypia. These X40 slides were labeled by 2 pathologists looking for individual instances of mitotic cells. When a cell was found, the center coordinate of that cell was used as the label along with a confidence score from 0.0 (non-mitosis) to 1.0 (true mitosis). When there was disagreement between the two pathologists on the state of a cell a third pathologist was made to examine the cell and the majority decision was adopted. In the end, the images that were used in the dataset were 1539×1376 pixels from the Aperio Scanscope XT and 1663x1485 from the Hamamatsu Nanozoomer scanner \cite{ICPR2014}.

\subsection{Data Pre-Processing}
 Some images in the original training set contained no instances of mitotic or non-mitotic cells in their labels so they were removed. In order to train a Faster-RCNN model the centroid labels given in the ground truth annotations were converted into 70x70 pixel bounding boxes. To speed up training the only data augmentation used in our pipeline was horizontal flipping. The data was then divided into a training set containing 1385 images and a validation set containing 347 images. During our initial training run there was an issue where the model was not producing bounding box proposals on the images. Our hypothesis was that the instances of cells in the images were too small for consistent detection by the model. To test out our theory we created 2 separate datasets made out of patches of the original images with varying patch sizes: 256x256 512x512, and 1024x1024 in order to get a larger representation of the cells in the images.

One major issue when it comes to quantitative analysis of histopathology slides is the variation in color as a result of the h\&e staining. In our pipeline, color normalization via the Reinhard method was implemented to combat the issue of color variation between stains. Reinhard's method involves matching the color distribution of one image to a target image with the use of a linear transformation in the $\iota\alpha\beta$ space defined in \cite{rutherford}. In our case $\iota$, $\alpha$, and $\beta$ are the red, green, and blue color channels of our image respectively. Reinhard's method was chosen because the transformation is simple and computationally cheap to implement via the following equations:

\begin{equation}
    \iota _{mapped} = \frac{\iota_{original} - \bar\iota_{original}}{ \hat{\iota}_{original}}\hat{\iota}_{target} + \bar\iota_{target}
\end{equation}

\begin{equation}
    \alpha _{mapped} = \frac{\alpha_{original} - \bar\alpha_{original}}{ \hat{\alpha}_{original}}\hat{\alpha}_{target} + \bar\alpha_{target}
\end{equation}

\begin{equation}
    \beta _{mapped} = \frac{\beta_{original} - \bar\beta_{original}}{ \hat{\beta}_{original}}\hat{\beta}_{target} + \bar\beta_{target}
\end{equation}

where $\bar\iota$, $\bar\alpha$, and $\bar\beta$ are the mean pixel activations of the RGB color channels respectively and 
$\hat{\iota}$, $\hat{\alpha}$, and $\hat{\beta}$ are the standard deviations of each channel of the image \cite{reinhard, colors}. Figure 2 shows an illustration of images that are transformed via the Reinhard method.

\subsection{Training and Inference}
The architecture used for training was a Faster-RCNN model using resnet-50 as the feature extractor. The initial learning rate used for our experiments was 0.005 with a cosine decay learning rate schedule implemented to combat overfitting. The optimizer used was Stochastic Gradient Descent with a momentum of 0.9. All models were trained for 25 epochs with early stopping implemented for 5 epochs to avoid overfitting.

\section{Results}

\subsection{Mean Average Precision}
The metric used for evaluation of our model was mean average precision (mAP). mAP is a commonly used metric in the field of object detection and has been used as the primary metric for evaluating models on major benchmark datasets like Microsoft COCO and Pascal VOC \cite{coco}. The calculation of mAP involves 3 major steps. The first step is find the area under the precision recall curve for every class in a dataset for a specific Intersection Over Union (IoU) threshold. The next step is to average the areas for every class together to get the Average Precision (AP) for one IoU threshold. The third and final step is to average all the APs together for IoUs $\in [0.5: 0.05:0.95]$(referred to as mAP@[0.5, .95] for MS COCO at mAP@0.5 for Pascal VOC \cite{ren}) to finally arrive at the mean average precision \cite{yohanandan}.

\begin{figure}
    \centering
    \includegraphics[width=5cm, height=5cm]{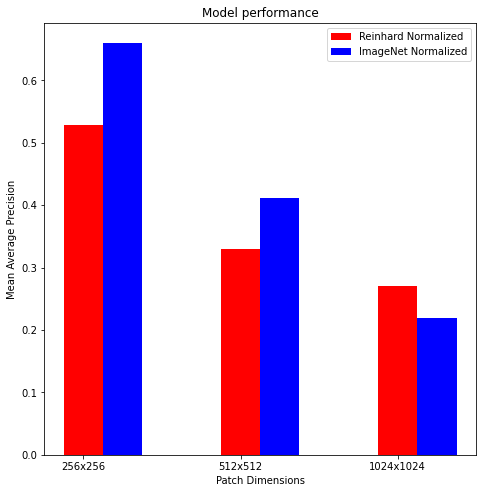}
    \caption{Chart detailing model performance with differing patch sizes as well as using Reinhard image normalization and the standard imagenet normalization included in the AutoML package.}
    \label{fig:my_label}
\end{figure}

\subsection{Validation Results}
After each model was trained it was evaluated on a validation dataset originally containing 357 images that were sliced into patches of the same size as the training patches. The mAP metric was evaluated using the standard pycocotools library to evaluate bounding box proposals. All the reported mAPs are using an IoU threshold of 0.5 similar to the Pascal VOC challenge. The reason for this choice is because the tool is designed to aid pathologists with localizing instances of mitotic cells, the predictions themselves do not need to be exact as long as the bounding box locates the approximate area of the cell for the pathologist to examine. The first model was trained with a patch size of 256x256 and no color normalization and achieved an mAP of 0.629. We then normalized all the images based on the pixel distribution from the first image in our training set and re-trained the model using these normalized images. This second model achieved an mAP of 0.528. Our next model was trained with patches that were 512x512 in size and the same process of training with unnormalized and normalized images was implemented once again. The mAPs were 0.329 and 0.411 for the unnormalized and normalized image models respectively. the final set of tests used patches of size 1024x1024 and achieved mAPs of 0.271 and 0.219 for the normalized and unnormalized images respectively. Figure 3 summarizes the results of our experiments with the different sets of hyperparameters used.

\section{Discussion}
\label{sec:disc}
The AutoML tool has performed well for the task of mitotic cell detection in histopathology images. It is apparent that performance seemed to decrease as the size of the individual patches increasd. One reason for this might the increasingly small representations of cells inside the images. It was brought up earlier in the paper that the AutoML tool would not produce any bounding box predictions for the original images due to the exceedingly small representations of the cells. This would help to explain why the model seemed to be performing worse as the patch sizes increased, the representation of cells inside the patches was becoming smaller and smaller thus harder to detect correctly. 

\subsection{Limitations and Future Work}
Despite the satisfactory performance of the AMDet tool, there are some limitations in this work and directions for future research. For one, only 3 dimensions were chosen for the patches during training. A future study could conduct a more thorough investigation to see how the AutoML tool fares with several larger or smaller image dimensions. Another weakness of this study is the normalization method used. The Reinhard method has been shown to help alleviate the issue of color variation in stains \cite{colors} however the method is over 2 decades old and there have been more recent advances in color normalization methods. Future work could examine the effect of using different color normalization methods in the AutoML package for mitotic cell detection. Furthermore, the AutoML tool contains standard ImageNet color normalization in the pipeline. This pre-built normalization was not removed during our experiments and seemed to have a negative effect on model performance as can be seen in Figure 3. Further studies can look at how the standard image normalization featured in the AutoML tool compares to different color normalization methods proposed.  

\section{Conclusion}
In this work AMDet, an automated tool for mitotic cell detection, was created with the use of the AutoML pipeline. To the best of our knowledge this is the first time the tool has been formally tested for detecting small objects within histopathology images. All the code is available at \href{https://github.com/WaltAFWilliams/AMDet}{\underline{ https://github.com/WaltAFWilliams/AMDet}}.

\bibliographystyle{unsrtnat}

\end{document}